\documentclass[conference]{IEEEtran}
\usepackage{authblk}

\usepackage{amsmath,amssymb,amsfonts}
\usepackage{algorithmic}
\usepackage{graphicx}
\usepackage{textcomp}
\usepackage{xcolor}
\usepackage[super,comma,sort&compress]  
   {natbib}
   
\title{Generalized multi-object classification and tracking with sparse feature resonator networks}

\author[1]{Lazar Supic}
\author[1]{Alec Mullen}
\author[1,2, 3]{E. Paxon Frady}
\affil[1]{Redwood Center for Theoretical Neuroscience, UC Berkeley}
\affil[2]{Neuromorphic Computing Lab, Intel}
\affil[3]{Correspondence: e.paxon.frady@intel.com}
\date{}
\begin{document}

\maketitle

\begin{abstract}
In many visual recognition and scene understanding tasks, it is essential to learn representations that capture both invariant and equivariant structure. 
While neural networks are frequently trained to achieve invariance to transformations such as translation, this often comes at the cost of losing access to equivariant information -- for example, the precise location of an object. 
Moreover, invariance to common transformations is not naturally guaranteed through supervised learning alone, and many architectures generalize poorly to input transformations not encountered during training.
Here, we take an approach based on analysis-by-synthesis and factoring using resonator networks. 
A generative model describes the construction of simple scenes containing MNIST digits and their transformations, like color and position. 
The resonator network can then be defined to invert the generative model, and provides both invariant and equivariant information about particular objects.
Sparse features learned from training data act as a basis set to provide flexibility in representing variable shapes of objects, allowing the resonator network to handle previously unseen digit shapes from the test set.
The modular structure of the resonator network provides a shape module which contains information about the object shape with translation factored out, allowing a simple classifier to operate on digits centered by the resonator network.
The classification layer is trained solely on centered data, requiring much less training data, and the network as a whole can identify objects with arbitrary translations without data augmentation.
The natural attention-like mechanism of the resonator network also allows for analysis of scenes with multiple objects, where the network dynamics selects and centers only one object at a time. 
Further, the specific position information of a particular object can be extracted from the translation module, and we show that the resonator can be designed to track multiple moving objects with precision of a few pixels.
\end{abstract}

\section{Introduction}
\label{sec:intro}

Our approach to analysis-by-synthesis \citep{renner2024neuromorphic} relies on a specification of a generative model and uses ``search in superposition'' with a recurrent network architecture called a resonator network \citep{frady_2020_resonator}.
The resonator network is a modular structure, where each module forms an estimate of one of the factors of variation that defines the generative model. 
Each factor of variation must be solved for simultaneously, and the resonator network will dynamically converge to a solution when a set of factors is found that explains the input. 

In this previous work, the input scenes were constructed of rigid fixed objects. 
Here, we extend the approach to handle objects that have variability in their shapes. 
Our input scenes are now generated using MNIST digits, which results in variability of each object class. 
We further separate training and testing images. 
For training, a sparse code of MNIST digit features is learned and used as a basis set for the shape module of the resonator network. 
Thus, rather than a single shape template being desired, we desire a sparse code that describes the variable object shape.
The resonator network's goal is then to factor overall object position from its shape represented by the sparse coefficients, allowing for downstream applications that can operate on both the invariant and equivariant properties of the object.

Here, we illustrate how the resonator network modules can be used to access both invariant and equivariant information of extracted objects.
This allows us to perform generalized translation-invariant object classification using extremely simple training methods. 
Further, we can access equivariant properties of the objects, and we show the network accurately tracking moving objects with pixel-level precision. 

\section{Methods}

\subsection{Extracting sparse features from the training set}

In order to handle the natural variability of MNIST digits, we developed a generative model to assume that objects in the scene are constructed from a basis set of sparse features, rather than rigid templates.  
The sparse features are learned from the training set of MNIST digits. In this procedure, we isolate the test and training samples from MNIST, $P_{train} \in \mathbb{R}^{60,000 \times 784}$ and $P_{test} \in \mathbb{R}^{10,000 \times 784}$. 
We examined two methods for establishing sparse features, PCA+ICA, which produces sparse features that are strictly orthogonal, and sparse dictionary learning, which is not constrained to be orthogonal.

For PCA+ICA, we start with PCA to decompose the MNIST digits, $\mathbf{P}_{train} = \mathbf{U} \mathbf{\Sigma} \mathbf{V}$. We include only the top 154 eigenvectors, and form the whitened and truncated training set $\acute{\mathbf{P}}_{train} = \mathbf{U} \mathbf{\Sigma}^* \mathbf{V}$, where $\mathbf{\Sigma}^*$ has diagonal value of 1 for included eigenvectors, and 0 everywhere else. We then use fast ICA to decompose the whitened data into the mixing matrix and sparse features: $\acute{\mathbf{P}}_{train}= \mathbf{M}\acute{\mathbf{F}}$, with $\acute{\mathbf{F}} \in \mathbb{R}^{784 \times 154}$. 
For sparse dictionary learning, we utilized the \emph{sporco} software library \citep{wohlberg2017sporco}, and extracted 500-784 basis functions, which can be considered as the matrix $\mathbf{F}$. Note that the sparse basis functions from fast ICA in $\acute{\mathbf{F}}$ are constrained to be orthogonal, while the basis functions derived from \emph{sporco} in $\mathbf{F}$ are not.

\subsection{Generative model of simple scenes and vector encoding}

The structure of the resonator network and the subsequent experiments are designed following a generative model of the input scene.
For this task, random scenes are generated and analyzed by the resonator network. 
First, 3 digits from the MNIST testing dataset are chosen randomly. The digits are placed in the scene with a random horizontal and vertical translation. 
The digits are also colored in one of 7 random colors. 
The 7 colors used in the generative model given by a matrix $\mathbf{B} \in \mathbb{R}^{3 \times 7}$ with, for instance, $\mathbf{B}_{cyan} = [0, 1, 1]$. In similar fashion as described above, the matrix $\mathbf{B}$ may be whitened and encoded as the matrix $\acute{\mathbf{B}}$.

The resonator network operates in a high-dimensional randomized vector space to utilize the ability to ``search in superposition'' for valid configurations of the generative model that explain the input scene.
The network must search over a large combinatorial space, and it uses a modular structure to form a guess about each factor of variation that is present. The network must search through the large space of all factor combinations to find a valid solution. 
The modular structure of the resonator network is defined based on inverting the generative model of the scene (see \citet{renner2024neuromorphic}). 
Each resonator module is responsible for estimating one of the factors of variation for an object in the scene, as defined by the generative model, and shares its estimate with the other modules.

First, critical to representing images is to use the \emph{vector function architecture} encoding of translation \citep{frady2022computing}. 
A particular translation is represented by the ``exponentiation trick'', where two randomized base vectors $\mathbf{h} \in \mathbb{C}^N$ and $\mathbf{v} \in \mathbb{C}^N$ correspond to horizontal and vertical translation respectively, with $N$ being the dimensionality. 
The pixel location, $(x, y)$, is represented by index vector $\mathbf{h}^x \odot \mathbf{v}^y$.
Any image (such as the sparse features or the input image) $Im(x,y)$ is encoded as a function over the pixel space via the superposition of index vectors weighted by their corresponding image pixel values $\mathbf{s} = \sum_{x,y} Im(x,y) \cdot \mathbf{h}^{x} \odot \mathbf{v}^{y}$. 
This image encoding has pivotal properties for enabling scene factorization. 
Most importantly, it ensures that the \emph{equivariant vector operation} for image translation is the binding operation, i.e. $\mathbf{s} \odot \mathbf{h}^{\Delta x} \odot \mathbf{v}^{\Delta y}$ is the representation of the image translated by $\Delta x, \Delta y$.

For notational convenience, we will denote the ``codebook matrix'' $\mathbf{\Phi} \in \mathbb{C}^{N \times M}$ as containing the full span of vectors that represent pixel locations of an image, where $M$ is the number of pixels in the image. The encoding of an image is then a matrix-vector product between the codebook matrix and the vectorized image: $\mathbf{\Phi} vec(Im) = \sum_{x,y} Im(x,y) \cdot \mathbf{h}^x \odot \mathbf{v}^y$. 
This is essentially multiplication of each pixel value with the ``VFA codevector'' that represents that pixel's location in the image.
For RGB images with a color dimension, we add randomized vectors to encode each color channel, and these vectors are stored in matrix $\mathbf{G} \in \mathbb{C}^{N \times 3}$. Here, a pixel is indexed by both its location and color channel. Color images thus have a codebook matrix that includes vectors for each pixel location and color channel, and the encoding is still described as a matrix-vector product between the codebook matrix and vectorized image: $\mathbf{\Phi} vec(Im) = \sum_{x,y,c} Im(x,y,c) \cdot \mathbf{h}^x \odot \mathbf{v}^y \odot \mathbf{G}_c$.

The resonator network is a recurrent architecture consisting of multiple resonator modules.
Each resonator module contains a codebook matrix, and receives information from other resonator modules. 
The other modules provide their current estimates for the other factors of variationthat define the features of each factor of variation. 
The shape module contains the sparse features learned from the MNIST training set encoded with $\mathbf{\Phi}$ into VSA space, $\acute{\mathbf{D}} = \mathbf{\Phi} \acute{\mathbf{F}}$, the color module contains features describing each possible coloration $\acute{\mathbf{C}} = \mathbf{G} \acute{\mathbf{B}}$, and the position modules contain the span of code vectors describing horizontal and vertical position, $\mathbf{H}_x = \mathbf{h}^x$ and $\mathbf{V}_y = \mathbf{v}^y$.

\subsection{Scene understanding inference pipeline}

\begin{figure*}[t]
  \centering
  \includegraphics[width=0.9\textwidth]{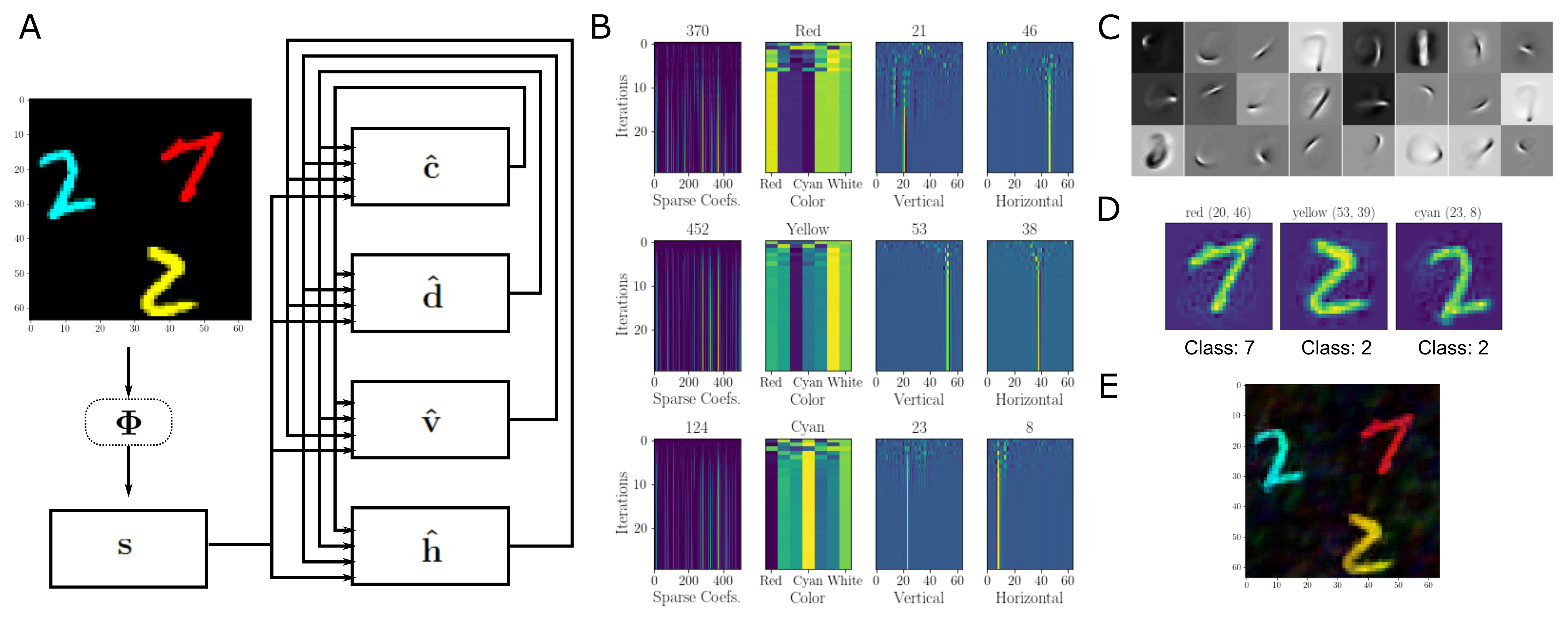}
  \caption{
  \textbf{Scene understanding with sparse feature resonator networks.}
  \textbf{A.} A simple scene with an MNIST digit is presented. The task is to factorize shape, location, and color, where each factor is represented by one of the resonator network modules.
  \textbf{B.} Visualization of the resonator network dynamics. Here, three resonator networks are operating in parallel, one for each row. Iteration time flows down. During early iterations dynamics are random an chaotic. Around iteration 5-10 the network finds a solution and converges. Yellow indicates highest output, the maximum peak is taken as the output for color and position. 
  \textbf{C.} Example sparse basis functions of MNIST digits were learned from a separate training set.
  \textbf{D.} The coefficients from the shape module and the sparse basis functions are combined to reconstruct the object with position and color factored out. A classifier then predicts digit identity from the centered digit; translation/color invariance is handled by the resonator network.
  \textbf{E.} The full scene is reconstructed from the resonator network outputs.
  }
  \label{fig:overview}
\end{figure*}

After the sparse basis functions are learned, we then are ready to perform visual scene understanding inference pipeline using the resonator network \citep{renner2024neuromorphic}.
The input image is first encoded into the high-dimensional vector space using the encoding matrix $\mathbf{s} = \mathbf{\Phi} vec(Im)$ (Fig. \ref{fig:overview}A). 

% \begin{figure}
%     \centering
%     \includegraphics[width=0.9\textwidth]{overview_figure-230914.png}
%     \caption{
%     \textbf{Sparse Feature Resonator Network}. 
%     A simple scene containing an MNIST digit is presented, and the task is to factorize shape, location and color. To do so a resonator network is constructed that contains modules that estimate each factor. B. When presented with a scene the recurrent resonator network iterates, searching for a factorization solution to the input problem. Each panel represents the state of a resonator module over time. At early iterations (top of each sub panel), the resonator network dynamics is chaotic. After several iteration, the network then finds a solution to the factorization problem and converges. C. Rather than storing templates in the resonator network as previously described \citet{renner2024neuromorphic}, sparse features of MNIST digits are learned from a separate training set. D. The sparse feature module of the resonator network extracts a sparse code that represents the object shape, here the sparse code for the object is reconstructed. The resonator network has factored out the object's location. E. Once the resonator network has factored out location, a classifier can be used to get the output digit identity. The classifier only needs to be trained on centered MNIST digits, as the resonator network takes care of the translation invariance.
%     }
%     \label{fig:overview}
% \end{figure}

The resonator network is a recurrent network that iteratively searches over factor configurations to find an explanation of the input image. 
An example of its dynamics is visualized in Figure \ref{fig:overview}B. The network first iterates searching chaotically over the configuration space (note upper regions of panels), until it finds a good match and rapidly converges to a solution (note consistent lines in lower regions of panels). Once it finds a good solution it remains stable and components describing an object can be interpreted from each module of the resonator network. 
The full dynamic equations for the conventional resonator network:
\begin{align}
\begin{split}
\mathbf{\hat{c}}(t+1) &= f \left( \mathbf{\acute{C}} \  g \left( \mathbf{\acute{C}}^\dagger \left( \mathbf{s} \odot \mathbf{\hat{d}}^*(t) \odot \mathbf{\hat{v}}^*(t) \odot \mathbf{\hat{h}}^*(t) \right) \right) \right), \\
\mathbf{\hat{d}}(t+1) &= f \left( \mathbf{\acute{D}} \ g \left( \mathbf{\acute{D}}^\dagger \left( \mathbf{s} \odot \mathbf{\hat{c}}^*(t) \odot \mathbf{\hat{v}}^*(t) \odot \mathbf{\hat{h}}^*(t) \right) \right) \right), \\
\mathbf{\hat{v}}(t+1) &= f \left( \mathbf{V}  \ g \left( \mathbf{V}^\dagger \left( \mathbf{s} \odot \mathbf{\hat{d}}^*(t) \odot \mathbf{\hat{c}}^*(t) \odot \mathbf{\hat{h}}^*(t) \right) \right) \right), \\
\mathbf{\hat{h}}(t+1) &= f \left( \mathbf{H}  \ g \left( \mathbf{H}^\dagger \left( \mathbf{s} \odot \mathbf{\hat{d}}^*(t) \odot \mathbf{\hat{v}}^*(t) \odot \mathbf{\hat{c}}^*(t) \right) \right) \right), \\
\end{split}
\label{eqn:resonator_transla}
\end{align}
with $f(\mathbf{x})_i = x_i/|x_i|$ (phasor projection) or $f(\mathbf{x})_i = x_i/||\mathbf{x}||_2$ (normalization). The function $g$ is either linear or an element-wise polynomial non-linearity.  

Here, we introduce an augmentation to the resonator network module based on including the \emph{locally competitive algorithm} (LCA) \citep{rozell2008sparse} as part of the dynamics. 
The LCA variant in particular has the ability to perform sparse inference with a non-orthogonal basis dictionary, due to its introduction of a competition term. 
Thus, the modules for shape and color can be implemented with the LCA dynamics and utilize unwhitened codebooks, $\mathbf{D} = \mathbf{\Phi} \mathbf{F}$ and $\mathbf{C} = \mathbf{G} \mathbf{B}$.
Since the codebooks for the position modules are naturally orthogonal, these still utilize the conventional dynamics as described above in Eq. \ref{eqn:resonator_transla}.

In the LCA resonator module, the activation coefficient of each dictionary basis function inhibits other activation values in proportion to the dot product of the basis functions. The activation coefficients are thresholded, so we define both internal state ($\mathbf{u(t)}$) and output activation coefficients ($\mathbf{x(t)}$). Here, we describe the shape module with codebook $\mathbf{D}$, and similar dynamics can be extended to the color module. The additional dynamic equations for the LCA resonator module are:
\begin{equation}
\begin{split}
\mathbf{u}(t+1) &= (1-\Delta)\mathbf{u}(t) \\
& + \Delta ( \mathbf{D}^\dagger (\mathbf{s} \odot \mathbf{o}^*(t)) - (\mathbf{D}^\dagger\mathbf{D} - \mathbf{I})\mathbf{x}(t)) \\
\mathbf{x}(t+1) &= T(\mathbf{u}(t+1); \lambda) \\
\mathbf{\hat{d}}(t+1) &= f\left(\mathbf{D}\mathbf{x}(t+1)\right)
\end{split}
\end{equation}
with $\Delta$ as a timestep parameter, and $T$ as a threshold function that is zero for inputs below threshold $\lambda$ and linear otherwise. 
The vector $\mathbf{o}=\mathbf{\hat{c}}(t) \odot \mathbf{\hat{v}}(t) \odot \mathbf{\hat{h}}(t)$ represents the factor estimates from other resonator modules.
Here, $-(\mathbf{D}^\dagger\mathbf{D} - \mathbf{I})\mathbf{x}(t)$ is the aforementioned competition term. 
Note that when $\Delta$ is small (we use $\Delta = 0.05$ in these experiments) there is a large contribution from the previous step, and this is helpful for LCA convergence. This hysteresis has been previously mentioned \citep{renner2024neuromorphic} and provides improvements in accuracy and stability. 

When there are multiple objects in the scene, the resonator network (typically) will ``attend'' to one object at a time. 
As described previously \citep{frady_2020_resonator}, the result can be explained away and the process repeated to analyze the next object, thus operating in \emph{serial}. Alternatively, multiple resonator networks can be instantiated and operate in \emph{parallel}. In the parallel mode, each network simultaneously explains away different aspects of the scene, enforcing that they attend to different objects.

Once the resonator network converges, the shape module now contains coefficients that represent the sparse code of the object being analyzed by the resonator network. 
Since the position and color modules have factored out the overall position and color of the object, the shape module can be decoded to reveal the centered and decolored object. 
The final converged state of the resonator network is then multiplied with the sparse basis functions (Fig. \ref{fig:overview}C) to recover the object shape $vec(\hat{Im}) = \mathbf{F}^\top \mathbf{D}^\dagger \mathbf{\hat{d}}$ (Fig. \ref{fig:overview}D). The outputs from all four factor modules and all three parallel resonator networks can be recombined to reconstruct the input scene (Fig. \ref{fig:overview}E).

All experiments with the resonator network were implemented in Python using NumPy and PyTorch. We explored two vector dimensions for the benchmarking experiments: N = 10,000 and N = 30,000.
%The preliminary analysis results showed no significant benefits of using N = 30,000, since the peak signal-to-noise ratio in the reconstructed images for N = 10,000 and N = 30,000 are within 0.5 dB. Conversely, a practical advantage of using N = 10,000 is that it reduces the running time needed for reconstruction by a factor of 3.
All experiments with the resonator network and different classifiers are done using the test MNIST dataset. 
%While all readers are familiar with the MNIST dataset, we would like to highlight several details of how this dataset is created. Usually, these details are overlooked by the community, but they are significant for our algorithm. 
%The MNIST dataset is a dataset that is a subset of a larger set available from NIST. 
The dataset consists of digits that have been size-normalized and centered in a fixed-size 28x28 pixel image.
%It is designed for individuals interested in learning techniques and pattern recognition methods using real-world data with minimal preprocessing and formatting efforts. 
%The original black and white images from NIST were adjusted to fit within the 20x20 pixel box while maintaining their aspect ratio, resulting in images with varying grey levels due to the anti-aliasing technique used during normalization. 
%The images were further centered within the 28x28 image by calculating the center of mass of the pixels and translating the image to position this point at the center of the 28x28 field. 
For our experiments, these images were placed in a larger (56x56)-(64x64) pixel image.

%To evaluate the resonator network model, the MNIST dataset is divided into two subsets, training and testing datasets. 
%The images are split into these two datasets using sklearn.model inbulit function test\_split with test\_size = 1/7. 
%After the split, the training dataset contains 60,000 images, and the testing dataset contains 10,000 images. 

\section{Results}

\subsection{Translation and color invariant patten classification}

The project's goal is twofold: 1.) to factorize a particular MNIST digit shape from color, and position. And, 2.) to classify correctly factorized MNIST digits. 
Once an MNIST digit's shape is found after factorization by the resonator network, the corresponding image is reconstructed from sparse feature components, with translation and color ``factored-out''. 
%Since the generative scene is 56x56 pixel image, the reconstructed image is 56x56  pixel image also. 
The reconstructed image is cropped to 28x28 image,  and fed as an input to a classifier. 
We used two classifiers: linear classifier and deep neural network (DNN) classifier. 
   
% and $\mathbf{V}$, $\mathbf{H}$ the codebooks of uncorrelated vectors representing vertical and horizontal coordinates of pixels. A linear transform of the form $\mathbf{V}\mathbf{V}^\dagger$ is essentially a linear auto-associative memory that aligns the output to the vectors closest to the input, stored in $\mathbf{V}$. 

% Classification results

% \begin{figure}
%     \centering
%     \includegraphics[width=0.9\textwidth]{Classification_results_complete.png}
%     \caption{Caption}
%     \label{fig:overview}
% \end{figure}

Summary of classification results on MNIST data sets is shown in table \ref{tab:mnist_accuracy}. 
The table presents classification accuracies of various models trained and tested on original and reconstructed MNIST data. 
%The highest accuracy (98.4\%) was achieved by the LeNet5 Neural Network trained and tested on the original MNIST dataset, reflecting the strong performance of convolutional architectures on clean input data. 
%When a Linear Classifier was used on the original MNIST data, it achieved a significantly lower but still respectable accuracy of 91.3\%, highlighting the performance gap between deep networks and simple linear models. 
While the resonator network factors out the color and location of the object, the output image is not a perfect reconstruction of the original digit. 
We trained the classifiers on both the original digits, as well as digits reconstructed by the resonator network.
The latter helps the classifier account for reconstruction artifacts due to the factorization procedure, and improves classification performance. 
We used both a simple Linear Classifier layer as well as LeNet5. 
Note that these classifiers are minimally trained with centered MNIST digits and there is no extra data augmentation.
%The Linear Classifier's accuracy dropped to 60.2\%. The LeNet5, although more robust, also suffered a performance drop, achieving 66.7\% accuracy. 
%Interestingly, training models directly on reconstructed MNIST images led to substantial gains in accuracy. 
The Linear Classifier II, trained and tested on reconstructed data, achieved 76.3\%. The LeNet5 trained and tested on reconstructed data reached 80.4\%. %indicating that end-to-end training on the altered image domain allows the models to partially adapt and recover performance.

\begin{table}[h]
\centering
\caption{Accuracy of Translation/Color Invariant Classifiers on MNIST scenes}
\begin{tabular}{|l|l|l|c|}
\hline
Classifier & Training Type & Accuracy \\
\hline
Linear I  & Original     & 60.2 \\
LeNet5 I  & Original    & 66.7 \\
Linear II & Reconstructed   & 76.3 \\
LeNet5 II & Reconstructed  & 80.4 \\
\hline
\end{tabular}
\label{tab:mnist_accuracy}
\end{table}

\subsection{Multi-object tracking}

Due to the modular structure of the resonator network, both object shape and position are simultaneously represented by different modules of the resonator network. 
The factorization procedure thus provides both invariant and equivariant information about object properties.

\begin{figure*}[t]
    \centering
    \includegraphics[width=0.9\linewidth]{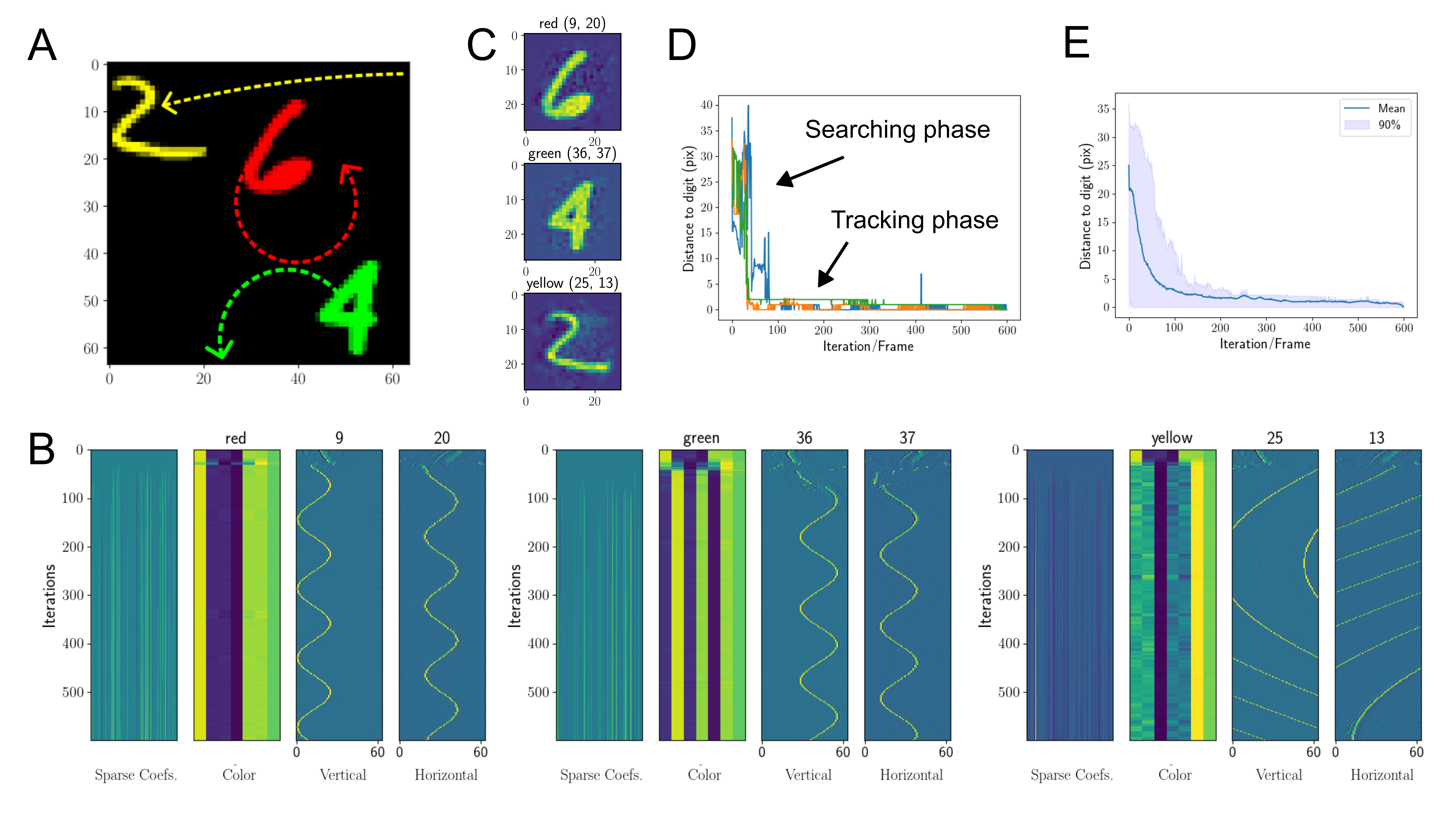}
    \caption{\textbf{Multi-object motion tracking with sparse feature resonator networks.}
    \textbf{A.} The input to the resonator network is a video of the simple MNIST scene with each digit moving along a trajectory, visualized by the colored arrows.
    \textbf{B.} The resonator dynamics are visualized during the tracking task. During initial iterations each network searches for one of the objects. Once an object is ``locked-on'', the shape and color modules converge, while the position module continues updating following the object. The object's trajectory can be decoded from the activity peaks of the position modules.
    \textbf{C.} The converged shapes of each resonator network's shape module is visualized.
    \textbf{D.} The distance between the estimated position and ground-truth position is calculated over time. The intial searching phase shows high errors, but once the object is locked on the tracking precision error drops to 1 or 2 pixels. 
    \textbf{E.} Summary of distance measurements from 100 3-object tracking experiments.
    }
    \label{fig:motion}
\end{figure*}

In the following experiments, we demonstrate multi-object motion tracking by the sparse feature resonator network. Here, three MNIST digits are present in the scene. Each digit in the scene is also moving along an independent straight, arc, or circular trajectory (Fig. \ref{fig:motion}A).
Note that objects in the scene will wrap around the scene borders, like a torus, and thus will never move out of view in these experiments.
Each frame from the video is encoded as before and presented to parallel sparse feature resonator networks monitoring the scene. 
In this experiment, the resonator network iteration rate is the same as the framerate of the movie.

As before, the resonator networks ``attend'' to one of the objects in the scene after an initial phase of searching for a valid factorization.
Once the resonator has locked-in on one of the objects, it continues to maintain a decomposition of the object's shape and position.
Since the object is moving, the position modules change over time, which can be see in the waterfall plots visualizing the resonator network dynamics (Fig. \ref{fig:motion}B, note the curvy yellow lines). We can further examine shape module to visualize the sparse features encoding of each MNIST digit (Fig. \ref{fig:motion}C).

The position modules can then be used to estimate the position of one of the objects in the scene. 
We measured the distance from the estimated position decoded from the resonator position modules with the ground truth distance over time (Fig. \ref{fig:motion}D).

Due to the fact that there is sometimes a consistent offset between the center of the object as determined by the resonator network, and the ground truth center, we include a small offset calibration to quantify the tracking behavior. This, in effect, calculates the precision of the resonator network tracking (Fig. \ref{fig:motion}E). 
The motion tracking of objects reaches 1-2 pixels of precision on average, and 95\% of objects are tracked to less than 5 pixels of precision. 
Without calibration the general accuracy of motion tracking is about 5 pixels, with 90\% of tracked objects have under 10 pixel distance of error.

\section{Discussion}

Here, we present a network architecture that can automatically disentangle location and color of a previously unseen object from its shape.
This approach to scene understanding starts with a generative model, where object configuration is specified by multiplicative binding of factors. 
The problem is then to find an appropriate configuration of the generative model that matches the input scene. 
The resonator network is a recurrent neural network architecture that solves this problem using a dynamic, iterative search strategy. 
By incorporating a sparse feature basis set into the shape module of the resonator network, we were able to extend beyond fixed object templates as described previously \citep{renner2024neuromorphic}. This then allows further downstream processing, like classification or motion tracking, based on factorized representations that contain both invariant and equivariant information about object properties.

Previous work \citep{kymn2024compositional} used convolutional sparse coding and resonator networks applied to the scene understanding task. 
This previous work applied convolutional sparse coding to the overall input scene, and the features for particular shapes in the scene are adjusted to the sparse basis set. 
This reduced the dimensionality and increased the sparsity of the factorization problem which improved the resonator network's performance, but still operated on template objects with rigid shape. 
Our effort here uses sparse features to develop a method for recognizing objects with variability in their shape by incorporating general sparse features into the shape module of the resonator network.
Fortunately, the approach of Kymn et al. \citep{kymn2024compositional} can be combined with our efforts to improve performance and efficiency with generalized object detection.

Unlike representation learning approaches to scene understanding \citep{bengio2013representation, higgins2017beta, locatello2020object}, our approach uses a-priori knowledge of the scene to formalize a generative model specified by \emph{multiplicative} combinations of object features.
This presents factorization as a new problem that needs to be solved. % in order to invert the generative model.
The advantage of this is that there is little learning required, and the system can automatically generalize to unseen objects in any pose.
The ability to generalize about unseen configurations of objects is known as ``out-of-distribution'' generalization \citep{locatello2019challenging, dittadi2021generalization, wiedemer2023provable}, and it is unlikely that supervised or semi-supervised learning approaches can truly achieve this without significant inductive biases \citep{fil2021beta, frady2023learning}. 
Factoring with the resonator network and factorized representation learning are promising approaches to enable out-of-distribution generalization in neural networks.

One issue with our current method is that the converged representation of the sparse code does not always result in object being exactly in the center.
There is sometimes a small offset from the true center, and this can interfere with both the classification performance and the tracking accuracy.
We attempted to build the classifier based solely on centered MNIST digits to avoid the use of data augmentation. However, minor data augmentation of small translations within 5 pixels may allow for much better classification performance. 
Alternatively, some extra mechanisms to encourage convergence to the true center could be helpful.

This approach to scene understanding uses the factorization to define both invariant and equivariant representations of the properties of objects. 
Notably, each resonator module is equivariant with the factor it is extracting, i.e. shape module will change in accordance with the shape of the object, and invariant to the other factors of variation, i.e. the shape module does not change in accordance with the object's position or color. 
The ability to factorize and access both invariant and equivariant representations allows for downstream applications targeting different properties. Further, this notably makes generalized classification problems much simpler, as variables irrelevant for classification are factored out. Thus, training can also be simplified, and the system naturally generalizes to unseen factor combinations. 

%Further uncovering the relationship between out-of-distribution generalization and factoring is critically important for understanding and improving neural networks.

\section{Acknowledgements}

We would like to thank Chris Kymn, Alpha Renner, Bruno Olshausen, Fritz Sommer, and members of the Redwood Center for Theoretical Neuroscience for feedback and discussion. 

\section{Contributions}

\textbf{Lazar Supic}: Conceputalization; Investigation; Formal analysis; Software; Writing -- review and editing; \textbf{Alec Mullen} Conceptualization; Investigation; Formal analysis; Software; Writing -- review and editing; \textbf{E. Paxon Frady}: Conceptualization; Supervision; Writing -- original draft preparation; Writing -- review and editing.

\bibliography{refs}

\end{document}